\documentclass{adobe_research}
\usepackage[T1]{fontenc}
\usepackage{lmodern}
\usepackage{multirow}
\usepackage{soul}
\usepackage{pifont}
\usepackage{graphicx}
\usepackage{amsmath,amssymb}
\usepackage{makecell}
\usepackage[dvipsnames]{xcolor}
\usepackage{color, colortbl}
\usepackage{caption}
\usepackage{algorithm}
\usepackage{newtxtext}
\usepackage{algpseudocode}
\usepackage{wrapfig}
\usepackage{float}
\definecolor{catgray}{gray}{0.92}

\usepackage[dvipsnames]{xcolor}
\usepackage{colortbl}
\usepackage{multirow}
\usepackage{pifont}
\usepackage{algorithm}
\usepackage{algorithmicx}
\usepackage{algpseudocode}
\usepackage{lineno}

\newcommand{\cmark}{\ding{51}}
\newcommand{\xmark}{\ding{55}}

\usepackage{xspace}
\makeatletter
\DeclareRobustCommand\onedot{\futurelet\@let@token\@onedot}
\def\@onedot{\ifx\@let@token.\else.\null\fi\xspace}

\makeatother

\newcommand{\ours}{SAR\xspace}

\title{Rethinking Training Dynamics in Scale-wise Autoregressive Generation}

\author[1*]{Gengze Zhou}
\author[2]{Chongjian Ge}
\author[2]{Hao Tan}
\author[2]{Feng Liu}
\author[2]{Yicong Hong}

\affiliation[1]{Australian Institute for Machine Learning, Adelaide University}
\affiliation[2]{Adobe Research}

\contribution[*]{Work done during internship at Adobe.}
\abstract{
\begin{abstract}

Recent advances in autoregressive (AR) generative models have produced increasingly powerful systems for media synthesis. Among them, next-scale prediction has emerged as a popular paradigm, where models generate images in a coarse-to-fine manner.
However, scale-wise AR models suffer from \textit{exposure bias}, which undermines generation quality. We identify two primary causes of this issue: (1) \textbf{train–test mismatch}, where the model must rely on its own imperfect predictions during inference, and (2) \textbf{imbalance in scale-wise learning difficulty}, where certain scales exhibit disproportionately higher optimization complexity.
Through a comprehensive analysis of training dynamics, we propose Self-Autoregressive Refinement (SAR) to address these limitations. SAR introduces a Stagger-Scale Rollout (SSR) mechanism that performs lightweight autoregressive rollouts to expose the model to its own intermediate predictions, thereby aligning train–test patterns, and a complementary Contrastive Student-Forcing Loss (CSFL) that provides adequate supervision for self-generated contexts to ensure stable training. Experimental results show that applying SAR to pretrained AR models consistently improves generation quality with minimal computational overhead. For instance, SAR yields a 5.2\% FID reduction on FlexVAR-d16 trained on ImageNet-256 within 10 epochs (5 hours on 32×A100 GPUs). Given its efficiency, scalability, and effectiveness, we expect SAR to serve as a reliable post-training method for visual autoregressive generation.

\end{abstract}
}

\date{December 1st, 2025}
\correspondence{\email{gengze.zhou@adelaide.edu.au, yhong@adobe.com}}

\adobedata[Project Page]{\url{https://gengzezhou.github.io/SAR}}

\begin{document}

\maketitle

\begin{figure}[h]
\begin{center}
   \includegraphics[width=0.98\linewidth]{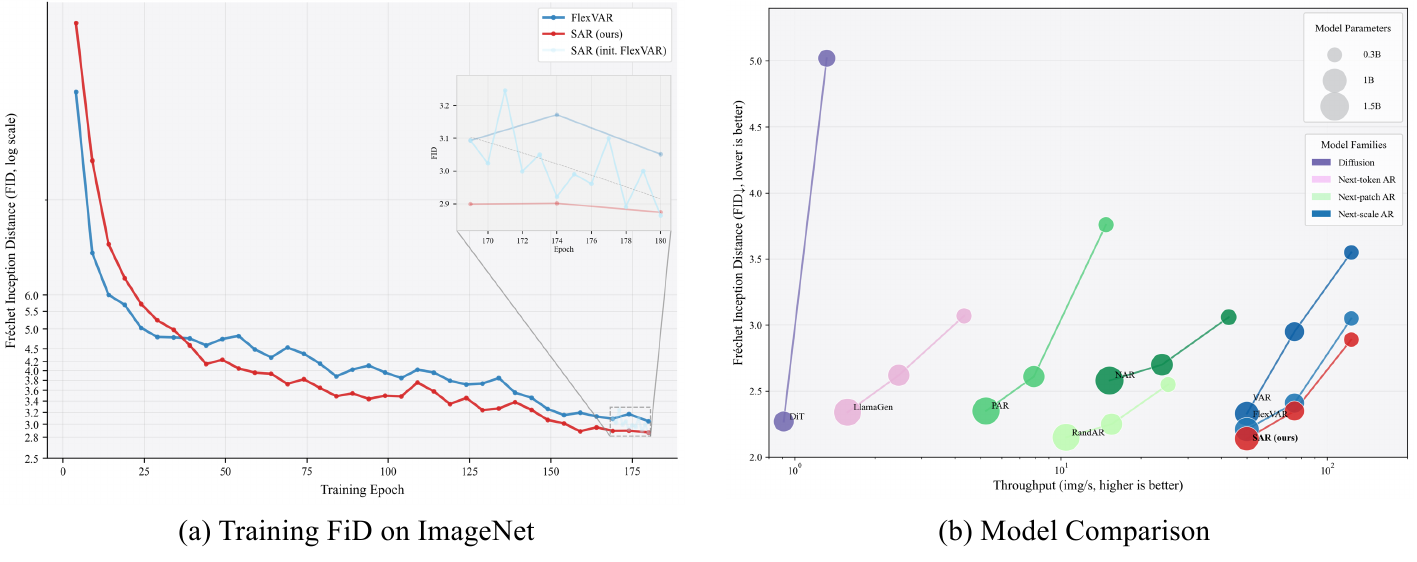}
\end{center}
\caption{
\textbf{(a) Training dynamics.} Training curves for FlexVAR, SAR trained from scratch, and SAR initialized from a pretrained FlexVAR checkpoint. Within only a few epochs, SAR quickly surpasses the best performance of a fully-trained FlexVAR model.
\textbf{(b) Throughput--FID trade-off.} Comparison of throughput, parameter count, and FID across representative generative model families, including Diffusion, Next-token AR, and Next-scale AR. \textbf{\ours} (red) attains the \emph{best overall trade-off}:  
the \textbf{highest throughput among autoregressive models} and further imporve next-scale prediction AR model with the \textbf{lowest FID across all AR baselines}.}
\label{fig:teaser}
\end{figure}

\section{Introduction}

\label{sec:intro}

Autoregressive (AR) modeling has achieved notable success in large language models (LLMs)~\citep{radford2018improving, radford2019language, brown2020language}, inspiring its adaptation to visual generation. By formulating image synthesis as a sequence modeling problem, AR models discretize images into ordered latent tokens and predict them causally. This formulation unifies autoregression across modalities and benefits from LLM-style scalability.
More recent progress in visual autoregressive generation (VAR)~\citep{VAR, han2024infinity} alleviates the limitation of spatial inductive bias by decomposing an image into a hierarchy of coarse-to-fine latent scales, predicting each scale conditioned on coarser representations. This next-scale-prediction formulation preserves spatial continuity, aligns better with the 2D nature of visual data, and significantly improves inference efficiency compared to token-level autoregression, whose sequence length grows quadratically with image resolution. 
However, existing VAR models suffer from two major issues, which negatively impact the generation quality: \textbf{(1) \textit{train–test mismatch}} and \textbf{(2) \textit{imbalanced learning difficulty across scales}}.

% the strictly causal generation order imposes a strong inductive bias: tokens generated early lack access to holistic semantic context, often leading to incoherent global structures and degraded spatial consistency.
% For these reasons, we build upon VAR as a scalable foundation for large-scale autoregressive visual generation. 

The first issue originates from \textit{exposure bias}, a persistent challenge inherited from autoregressive models. Existing VAR models are trained with \textit{teacher forcing}, i.e., conditioning on early-scale latents along the ground-truth trajectory. During inference, however, the model must rely on its own imperfect predictions, resulting in distributional drift and progressively accumulating errors during next-(group-of-)token prediction. This problem is more pronounced in VAR models than in token-wise autoregression, as each step involves predicting a large number of tokens at a scale. Errors at each scale propagate through the hierarchy, significantly degrading the final output quality.

The scale-wise prediction leads to the second issue, \textit{imbalanced learning difficulty across scales}. As will be shown in \S3.3, a typical pattern is that early coarse scales are responsible for generating global structure from scratch, whereas later fine scales mainly reconstruct high-frequency details under increasingly certain conditions, similar to spatial super-resolution. Minor inaccuracies at coarser scales are propagated and amplified through hierarchical decoding, causing cumulative errors and loss of visual fidelity. 
Despite this asymmetry, conventional supervision is typically applied uniformly across all scales, which means that once the coarse structure is formed in the early stages, the finer scales lack the capacity to correct structural artifacts introduced earlier. Consequently, training becomes uneven, and fine-scale refinement remains suboptimal.

In this paper, we first conduct a comprehensive empirical study to reveal the limitations of existing VAR formulations and the challenges of naively integrating student-forcing into training. Build on the emperical findings, we obtanin inspiration from the student-forcing formulation in robotic navigation~\citep{anderson2018vln,zhou2024navgpt2} and video generation~\citep{lin2025apt2,huang2025self}, and introduce Self-Autoregressive Refinement (SAR), a lightweight post-training algorithm that explicitly unrolls autoregressive inference during training to address these challenges. SAR bridges the train–test gap by conditioning each scale’s prediction on \textit{self-generated} latents rather than ground-truth inputs, directly exposing the model to inference-time uncertainty and balance scale-wise learning difficulty. Importantly, the scale-wise autoregressive structure of VAR makes student-forcing computationally feasible: a single model forward pass yields a full set of tokens for an entire scale, significantly reducing rollout cost.

Specifically, SAR introduces two key components: a Stagger-Scale Rollout (SSR) strategy and a Contrastive Student-Forcing Loss (CSFL). SSR performs parallelized single-step student rollouts based on teacher-forced predictions, enabling efficient exposure to self-generated latents with minimal computational overhead. CSFL explicitly aligns the outcomes of student and teacher rollouts, providing targeted supervision that stabilizes training under student-forcing conditions. Together, these components allow the model to adapt effectively to inference-time dynamics with only one additional forward pass.
In addition, SAR dynamically reweights per-scale supervision to address gradient sparsity, being effective in balacing optimization between coarse and fine scales and enhancing overall generation quality.

% and a \textbf{Contrastive Student-Forcing Loss (CSFL)} that explicitly aligns the outcomes of student and teacher rollouts. Together, these techniques stabilize training by providing adequate supervision across scales and enable the model to adapt to self-generated rollouts with only one additional forward pass.
% Lastly, SAR also dynamically reweights per-scale supervision to mitigate gradient sparsity and achieve balanced optimization between coarse and fine scales.

Experimental results show that applying Scale-wise Autoregressive Refinement (SAR) to pretrained state-of-the-art VAR models yields consistent improvements of 5.2\% / 2.5\% / 3.1\% lower FID on 300M, 600M, and 1B parameter models respectively, while using only 160 A100 GPU hours (32 80~GB A100s, ImageNet-256 for 10 epochs, amounting to $\approx$5.5\% of pretraining cost). These results demonstrate that SAR effectively reduces exposure bias, mitigates scale-wise learning imbalance, and enhances perceptual quality and robustness in large-scale visual autoregressive generation.

\section{Related Work}
\label{sec:related}

\noindent\textbf{Vector-Quantized Latent Modeling.}
VQ-VAE~\citep{vqvae} established a pioneering two-stage paradigm for discrete image generation: first encoding images into a latent space and quantizing them with a fixed-size codebook, and then modeling the discrete indices via an autoregressive prior such as PixelCNN~\citep{pixelcnn}. This formulation bridges continuous visual data with discrete sequence modeling, laying the foundation for modern visual autoregressive (AR) generation.

\smallskip
\noindent\textbf{Raster-scan Autoregression.}
Following this paradigm, subsequent works~\citep{esser2021taming, ramesh2021zero} employ Transformer-based AR priors over quantized latents. Hierarchical extensions such as VQ-VAE-2~\citep{vqvae2} and RQ-Transformer~\citep{lee2022autoregressive} introduce multi-scale or stacked latent codes to improve global coherence.  
Recent large-scale models~\citep{llamagen, liu2024lumina} adopt GPT-style next-token prediction to achieve high-fidelity image synthesis, while MARS~\citep{he2024mars} integrates a mixture-of-experts design to enhance diversity. AIM~\citep{aim} further incorporates the Mamba architecture~\citep{gu2023mamba} to accelerate sequential decoding.  
To mitigate the quantization-induced information loss, several approaches combine diffusion or denoising objectives with AR decoding~\citep{xie2024show, zhou2024transfusion, gu2024dart}, achieving improved detail reconstruction and perceptual quality.

\smallskip
\noindent\textbf{Random-scan Generation.}
In contrast to rasterized decoding, masked-prediction models~\citep{devlin2018bert, he2022masked, bao2021beit} adopt a BERT-style bidirectional formulation, predicting masked tokens conditioned on visible ones. This enables non-sequential token inference and parallel generation.  
MaskGIT~\citep{chang2022maskgit} and Muse~\citep{chang2023muse} extend this paradigm to class-conditional and text-to-image synthesis, achieving remarkable efficiency–quality trade-offs.  
The MagViT series~\citep{yu2023magvit, luo2024open} generalizes masked prediction to spatiotemporal domains, while NOVA~\citep{deng2024autoregressive} introduces a hierarchical decoding order—first over time, then over spatial token groups—to produce coherent video frames.

\smallskip
\noindent\textbf{Scale-wise Generation.}
Visual Autoregressive (VAR) modeling~\citep{VAR} redefines the decoding order from next-token prediction to next-scale prediction, generating all tokens at a given resolution in parallel. This substantially reduces inference steps while preserving global consistency.  
Subsequent advances such as VAR-CLIP~\citep{zhang2024var} extend this paradigm to text-to-image generation via CLIP-based conditioning. Efficiency-oriented improvements include fast decoding~\citep{code}, linear attention acceleration~\citep{ren2024m}, and lightweight quantizers~\citep{li2024imagefolder}.  
To enhance fidelity, HART~\citep{tang2024hart} and FlowAR~\citep{ren2024flowar} combine continuous tokenizers with flow matching or diffusion objectives, while Infinity~\citep{han2024infinity} reformulates AR prediction at the bitwise level, achieving unprecedented visual detail and scalability. However, current VAR architectures provide no mechanism to correct accumulated multi-scale errors, since training is strictly teacher-forced and offers no exposure to model-generated trajectories. This mismatch underscores the need for a more stable and computationally efficient autoregressive refinement strategy that is robust to error accumulation.

\section{Preliminaries}

\subsection{Scale-wise Autoregressive Visual Generation}

\noindent\textbf{Formulation.}
Scale-wise autoregressive (AR) models decompose image generation into a sequence of progressively refined visual scales, where each scale is conditioned on all preceding ones. Let the input image be represented by multi-scale latent maps $\{f_1, f_2, \dots, f_n\}$, each at an increasingly higher resolution $h_i \times w_i$. The model first predicts the smallest-scale latent $f_1$, and then autoregressively generates the subsequent latents conditioned on all earlier scales:
\begin{align}
    p(f_1, f_2, \dots, f_n) = \prod_{i=1}^{n} p(f_i \mid f_{1}, f_{2}, \dots, f_{i-1}).
\end{align}
Each $f_i$ may represent either a residual latent (as in residual VAR~\citep{VAR, zhang2024var, tang2024hart}) or a ground-truth latent (as in FlexVAR~\citep{jiao2025flexvar, yu2025frequency, ren2024flowar}).  
Residual-based methods define $f_i = g_i - \mathrm{Upsample}(g_{i-1})$, emphasizing difference modeling, while ground-truth-based models directly predict $f_i = g_i$.  
Both paradigms employ attention-based architectures (e.g., Transformers~\citep{vaswani2017attention}) to capture long-range dependencies across scales.  
Unlike token-level autoregression, where one token is predicted per step, scale-wise AR generates an entire latent map per step, preserving the 2D structure and greatly improving efficiency.

\noindent\textbf{Tokenization Process.}
Before autoregressive modeling, the input image $I$ is encoded into hierarchical discrete latent representations through a multi-scale tokenizer.  
Specifically, an image tokenizer (e.g., VQ-VAE or residual quantizer~\citep{lee2022autoregressive, code}) maps $I$ to a set of latent feature maps $\{f_1, f_2, \dots, f_n\}$ with resolutions $\{h_1, h_2, \dots, h_n\}$.
Each latent map $f_i$ can be further quantized into discrete code indices $t_i \in \mathcal{Z}^{h_i \times w_i}$, forming the multi-scale token maps $T = \{t_1, t_2, \dots, t_n\}$.
During autoregressive rollout, the model sequentially generates $t_i$ (or its continuous embedding $f_i$) conditioned on all previously generated scales.
Depending on the tokenizer design, $f_i$ can represent either an \textit{upsampling pathway} (top-down prediction)~\citep{ren2024flowar,jiao2025flexvar} or a \textit{residual pathway} (difference-based refinement)~\citep{VAR,han2024infinity}, offering flexibility in how hierarchical information propagates across scales. In this work, we primarily investigate the upsampling pathway for its simplicity and ease of formulation and training, while achieving comparable performance to residual-based alternatives.

% \noindent\textbf{Flexible Positional Embedding.}
% Traditional VAR models use fixed-length positional embeddings (PEs) tied to a predefined number of scales and resolutions, which constrains both training and inference.  
% To enable arbitrary scale configurations, FlexVAR introduces a scalable 2D positional embedding $\mathcal{P} \in \mathbb{R}^{d \times 2h \times 2w}$ initialized with a 2D sine–cosine pattern~\citep{vit}.  
% At the $i^{th}$ autoregressive step, $\mathcal{P}$ is adaptively resized via bilinear interpolation to match the spatial size of $f_i$.  
% This design preserves positional consistency across scales while allowing flexible addition or removal of autoregressive steps.  
% Empirically, we find that explicit step embeddings—commonly used to indicate generation order—are unnecessary under the ground-truth prediction paradigm.  
% Removing them improves generalization and enables continuous scaling during both training and inference.

\subsection{Teacher Forcing in Scale-wise AR}
VAR and its successors typically follow a teacher-forcing (TF) training paradigm: at each scale $i$, the model receives the \emph{ground-truth} latent $f_{1:i-1}$ and predicts the next-scale latent (GT or residual) $f_i$. Formally,
\begin{equation}
    \hat f_i = g_\theta(f_{1:i-1}), \qquad 
    \mathcal{L}_{\text{TF}} = \sum_{i=1}^{N} \ell(\hat f_i, f_i),
\end{equation}
where $\ell$ is an $L_2$ or cross-entropy loss depending on the tokenizer.
To construct inputs for all scales in parallel (analogous to the next-token prediction in LLMs), we precompute
$f_{1}, f_{2}, \dots, f_{N}$
and feed them to the model in one forward pass, with “scale shift” aligning each scale $f_{i}$ to predict $f_{i+1}$.

This TF strategy is fully parallelizable and efficient, but induces a severe \textbf{train–test gap}: inference must generate all scales autoregressively using its \emph{own} predictions, but training never exposes the model to such self-generated inputs.
This motivates the use of \emph{student-forcing} to reduce the discrepancy between training and inference.

\begin{figure}[t]
\begin{center}
   \includegraphics[width=0.99\linewidth]{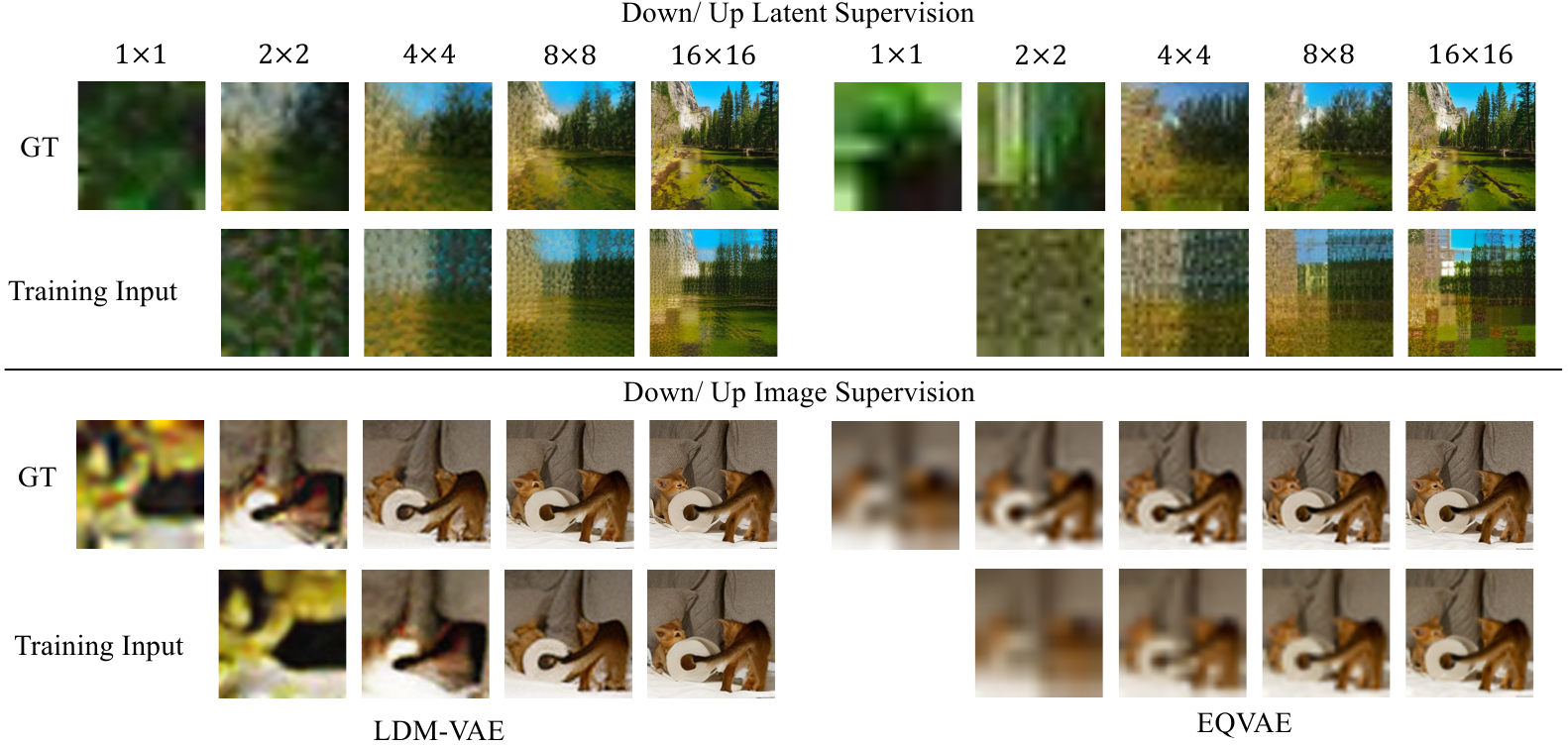}
\end{center}
   \caption{
    \textbf{Illustration of training supervision imbalance.} For \textbf{latent-space (top) supervision}, coarse scales receive ground-truth signals that contain little semantic structure, while their corresponding training inputs are dominated by blurry upsampled artifacts. Consequently, the finest scale must reconstruct nearly all details, causing the hierarchical prediction process to collapse into a single dominant scale and preventing effective coarse-to-fine learning. We could smooth the generation trajectory by \textbf{downsampling in image space (bottom)}, but this causes the earliest scales to capture most semantics and scene structure already, leaving later scales to perform only mild sharpening or super-resolution, and thereby weakening the multiscale factorization of the model.}
\label{fig:imbalance}
\end{figure}

\subsection{Diagonosing Scale-wise Autoregression}
\label{sec:imbalance}

\begin{figure}[t]
\begin{center}
   \includegraphics[width=0.99\linewidth]{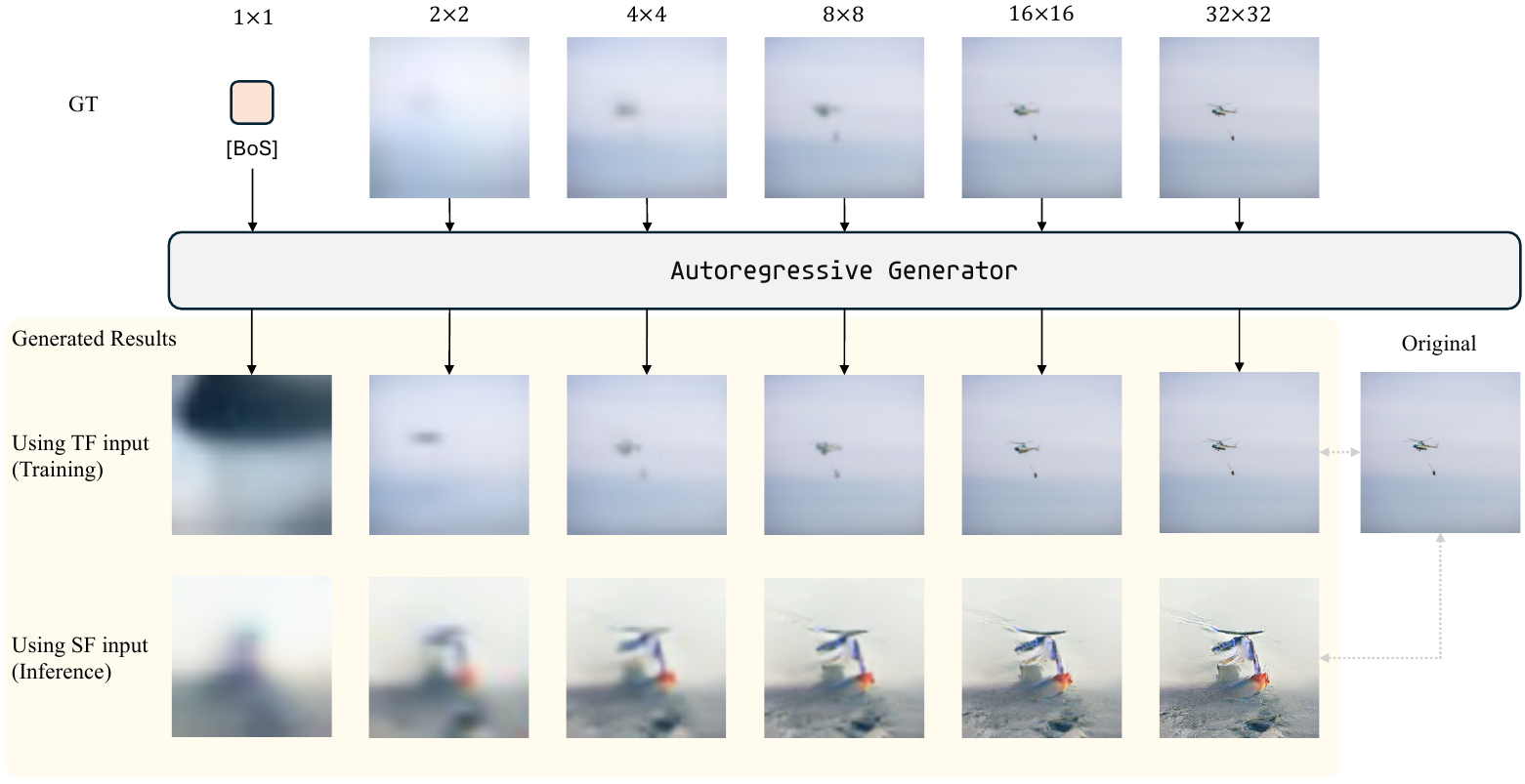}
\end{center}
\caption{
\textbf{Training–inference divergence caused by scale-wise supervision imbalance (Sec.~\ref{sec:imbalance}).}
Under teacher-forcing (top), the model receives ground-truth latents at all scales; when training converged, the model produces clean generated results because it is evaluated under the same idealized inputs used during training. 
At inference (bottom), the model must condition on its own coarse-scale predictions. 
When generated early-scale latents are imperfect (e.g., 1$\times$1), later scales, which were trained mainly as super-resolution task when using the smoothed up/down-sample image supervision, cannot correct the semantic error, leading to a complete collapse of the generation process.
}

\label{fig:inference_imbalance}
\end{figure}

We investigate the generation behavior of scale-wise autoregressive (VAR) models and uncover a critical limitation.
Despite their elegant hierarchical formulation of predicting finer-scale latents conditioned on coarser ones, training such models in practice leads to a severe \textit{scale-wise supervision imbalance}. As illustrated in Fig.~\ref{fig:imbalance} (top), under the conventional \textit{down/up latent supervision} scheme, where ground-truth latents at each scale are obtained by downsampling the full-resolution latent, and inputs are produced by upsampling the ground-truth from the previous scale—the early coarse scales receive supervision devoid of meaningful semantic structure, and the final fine scale bears the majority of the reconstruction burden, having to refine extremely blurry and blocky inputs into detailed images.
This imbalance effectively collapses the intended coarse-to-fine hierarchy into a single dominant scale, undermining both the efficiency and interpretability of the autoregressive generation process.  

To systematically analyze this issue, we conducted a series of studies aimed at disentangling potential causes. Specifically, we evaluate whether the imbalance can be alleviated by: (1) smoothing the supervision path through \textit{down/up-sample image supervision} or (2) replacing the VAE with a scale-equivariant VAE, and (3) introducing hybrid masked–autoregressive modeling to enable more NFE to early-scale generation.

\begin{wraptable}{r}{0.55\linewidth}
% \begin{table}[t]
\footnotesize
\centering
\caption{\textbf{Generation performance on ImageNet 256$\times$256.}
Comparison of image- vs.\ latent-level supervision for continuous VAEs paired with FlowAR-B. Metrics: FID, IS, precision (Pre), resolution (Res).}

\resizebox{\linewidth}{!}{
\begin{tabular}{lcc|cccc}
\toprule
{VAE} & {AR Model} & {Supervision} & {FID$\downarrow$} & {IS$\uparrow$} & {Pre$\uparrow$} & {Res$\uparrow$}\\
\midrule
\rowcolor{adobebg}\multicolumn{7}{l}{\emph{Continuous VAE}:}\\
\multirow{2}{*}{LDM-VAE} &  \text{FlowAR-B}  &  \text{Image} &  7.48   &  390.3   &  0.88   &   0.36  \\
 &  \text{FlowAR-B} &  \text{Latent} &  3.90   &  232.6   &  0.80   &   0.51  \\
\midrule
\multirow{2}{*}{EQVAE} &  \text{FlowAR-B} &  \text{Image} &  7.89   &  155.8   &  0.80   &   0.43   \\
&  \text{FlowAR-B} &  \text{Latent} &  3.93   &  233.4   &  0.81   &   0.50  \\
% \midrule
% \rowcolor{Cerulean!20}\multicolumn{7}{l}{\emph{Discrete VAE}:}\\
% \multirow{2}{*}{VAR-VAE} &  \text{FlowAR-B} &  \text{w/ Quant.} &  5.24   &  206.0   &  0.83   &   0.44  \\
% &  \text{FlowAR-B} &  \text{w/o Quant.} &  4.98   &  209.3   &  0.81   &   0.46   \\
\bottomrule
\end{tabular}
}
\vspace{-10pt}
\label{tab:gen_vae}
% \end{table}
\end{wraptable}

\noindent\textbf{Smoothing Scale-wise Supervision.}  
We first compare latent-level and image-level supervision under two different VAEs: a standard latent diffusion VAE (LDM-VAE) and an equivariant VAE (EQVAE) trained jointly on up/down-sample image and latent generation. The experiments use FlowAR-B~\citep{ren2024flowar}, a scale-wise AR model with a diffusion head for regressing continuous latents, trained for 50 epochs on ImageNet 256×256.  
As shown in Table~\ref{tab:gen_vae}, image-level supervision produces visually smoother input–target pairs but results in substantially higher FID scores (7.5–7.9 vs.\ 3.9), indicating that overly smooth latent supervision is ineffective. The EQVAE tokenizer facilitates even smoother feature propagation from the initial scale, the generation performance still lags behind. 

To further illustrate the consequences of \emph{scale-wise supervision imbalance}, we visualize the intermediate predictions produced during training (teacher-forcing) and inference (student-forcing), as shown in Fig.~\ref{fig:inference_imbalance}. When provided with ground-truth inputs at every scale during training, the model exhibits a stable coarse-to-fine refinement trajectory: early scales recover the global structure, and subsequent scales progressively sharpen local details, ultimately reconstructing an image nearly indistinguishable from the original. This confirms that the model has indeed learned to perform effective multi-scale denoising under ideal (teacher-forced) conditions.

In contrast, during inference the model must autoregressively condition on its \emph{own} predictions. If the generated earliest coarse-scale latent (e.g., the 1$\times$1 scale initialized only from a \textsc{[bos]} token) is bad, the subsequent scales fail to recover meaningful semantics. This degradation stems directly from the supervision imbalance discussed earlier: later scales are predominantly trained to perform super-resolution on \emph{ground-truth} upsampled latents, rather than to correct semantically corrupted coarse predictions. As a result, the generation process collapses as later scales faithfully magnify erroneous early predictions instead of repairing them, leading to severely distorted final outputs despite the model being well-trained under teacher-forcing.

This visualization therefore makes explicit the mismatch between training and inference dynamics: training on latent down/up-sampling makes later-scale supervision disproportionately challenging, while training on image down/up-sampling makes later-scale supervision insufficiently informative. Consequently, inference is disproportionately sensitive to coarse-scale errors, revealing a fundamental brittleness in conventional scale-wise autoregressive generation.

% \begin{table}[t]
\begin{wraptable}{r}{0.55\linewidth}
\centering
{
\caption{\textbf{Hybrid modeling performance on ImageNet 256$\times$256.}
Applying MaskGIT-style prediction at the coarsest scales strengthens initial 4$\times$4 representations, but leads to degraded full-resolution FID, indicating that later scales collapse into near super-resolution refinement rather than benefiting from scale-wise autoregression.}
\vspace{-1em}
\begin{center}
   \includegraphics[width=0.85\linewidth]{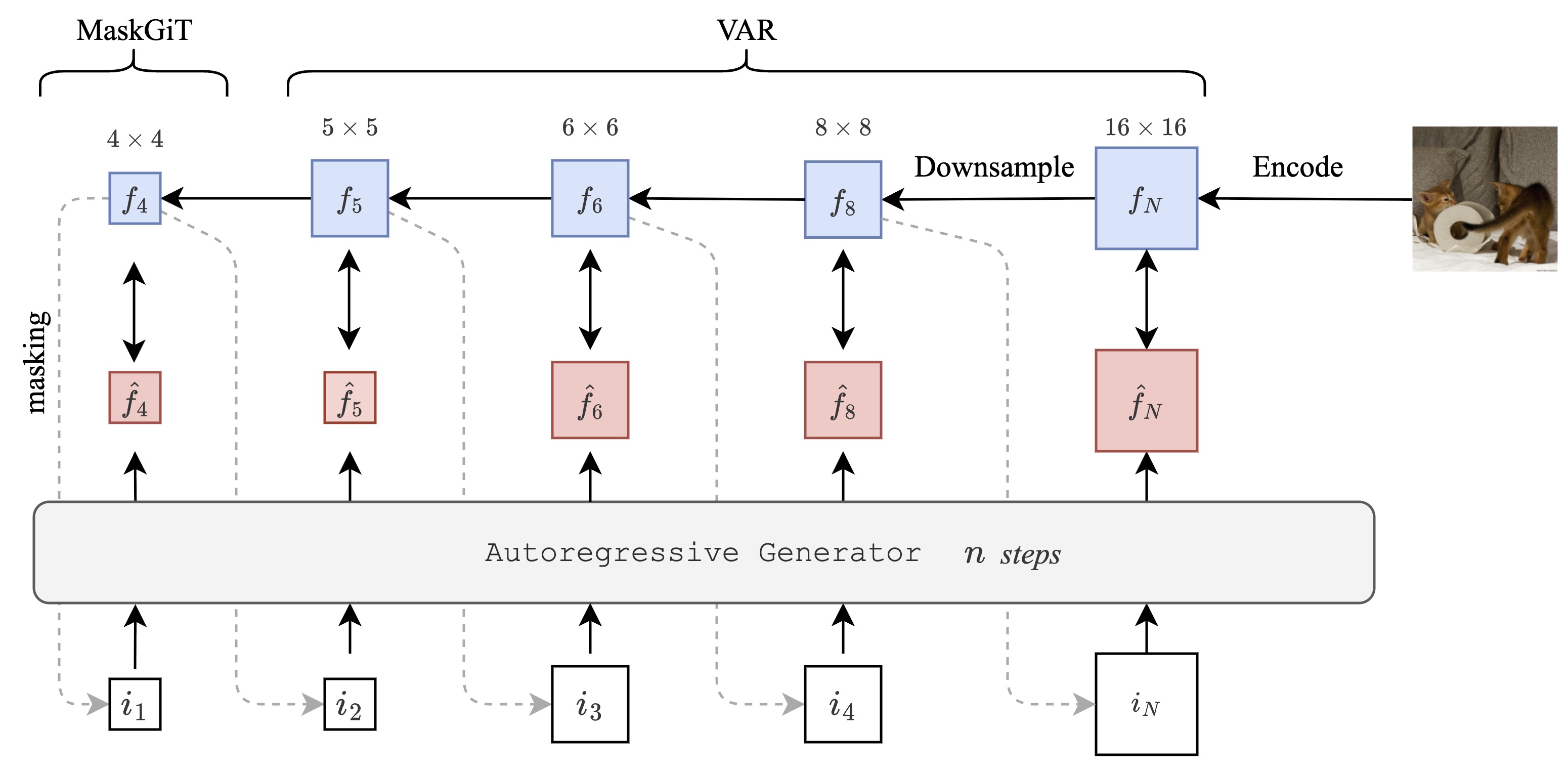}
\end{center}
\label{tab:hybird}
}
\vspace{0.5em}
\resizebox{\linewidth}{!}{
\small
\begin{tabular}{l|c|cc|cc}
\toprule
 Model         & FID @ 4 & FID$\downarrow$ & IS$\uparrow$ & Pre$\uparrow$ & Rec$\uparrow$  \\
\midrule
\rowcolor{adobebg}\multicolumn{6}{l}{\emph{Discrete VAE}:}\\
FlexVAR-$d16$      & 27 & 3.05  & 291.3 & 0.83&0.52   \\
\rowcolor{Cerulean!10}
\quad w/ Hybird Modeling & 22 & 5.02  & 250.2 & 0.81 & 0.47   \\

FlexVAR-$d20$    & 26  & 2.41  & 299.3 & 0.85 & 0.58  \\
\rowcolor{Cerulean!10}
\quad w/ Hybird Modeling & 20 & 4.74  & 303.6 & 0.84 & 0.44   \\

\bottomrule
\end{tabular}
}
% \begin{figure}[h]
% \begin{wraptable}{r}{0.4\linewidth}

   % \caption{
    % Illustration of hybrid AR modeling.}
\label{fig:hybird}
% \end{figure}
% \end{wraptable}

% \end{table}

\end{wraptable}

\noindent\textbf{Decompose Initial-scale Generation.}  
To further address the supervision imbalance, we design a hybrid training scheme that applies masked-token modeling (MaskGIT-style) at the early scales to enhance global structure learning, followed by scale-wise autoregression at higher resolutions. The key idea is to decompose the challenging initial-scale generation into multiple iterative steps, a strategy shown effective in prior work~\citep{li2024mar,pang2025randar}. We instantiate this hybrid model on FlexVAR and apply \textit{down/up-sample latent supervision}, enabling direct comparison against standard FlexVAR variants.
As shown in Table \ref{tab:hybird}, the hybrid modeling indeed improves FID at the coarsest 4×4 scale (from 27 to 22 for $d{=}16$, and form 26 to 20 for $d{=}20$), demonstrating stronger coarse-scale representation. However, the overall FID of full-resolution generation degrades notably (from 3.05 to 5.02 and from 2.41 to 4.74), accompanied by reductions in both recall and precision. 
These results indicate that, although early scales benefit from bidirectional mask modeling, the later scales collapse into near super-resolution refinement, preventing the model from fully exploiting the intended scale-wise autoregressive dependencies.

\vspace{4pt}
\noindent\textbf{Findings.}
Collectively, these results suggest that neither smoothing the supervision path (latent vs.\ image) nor allocating additional computation to the early scales effectively resolves the imbalance. We hypothesize that both strategies shift the modeling burden to the earliest scale, where the model must create the entire global structure from scratch, while later scales become overly simplified, functioning merely as super-resolution that are easy to optimize and prone to overfitting. 
Overall, these analyses demonstrate that scale-wise autoregressive frameworks require a fundamental rethinking of their training objectives and dependency structures, beyond residual formulations or supervision design, to achieve balanced hierarchical generation.

\section{Self-Autoregressive Refinement}
We propose applying \textit{Student Forcing} training to the autoregressive process, which is widely used to mitigate the training–inference discrepancy. In our case, it also serves to balance scale-wise learning difficulty by propagating early-scale imperfections to later scales.

\begin{figure*}[ht]
\begin{center}
   \includegraphics[width=0.99\textwidth]{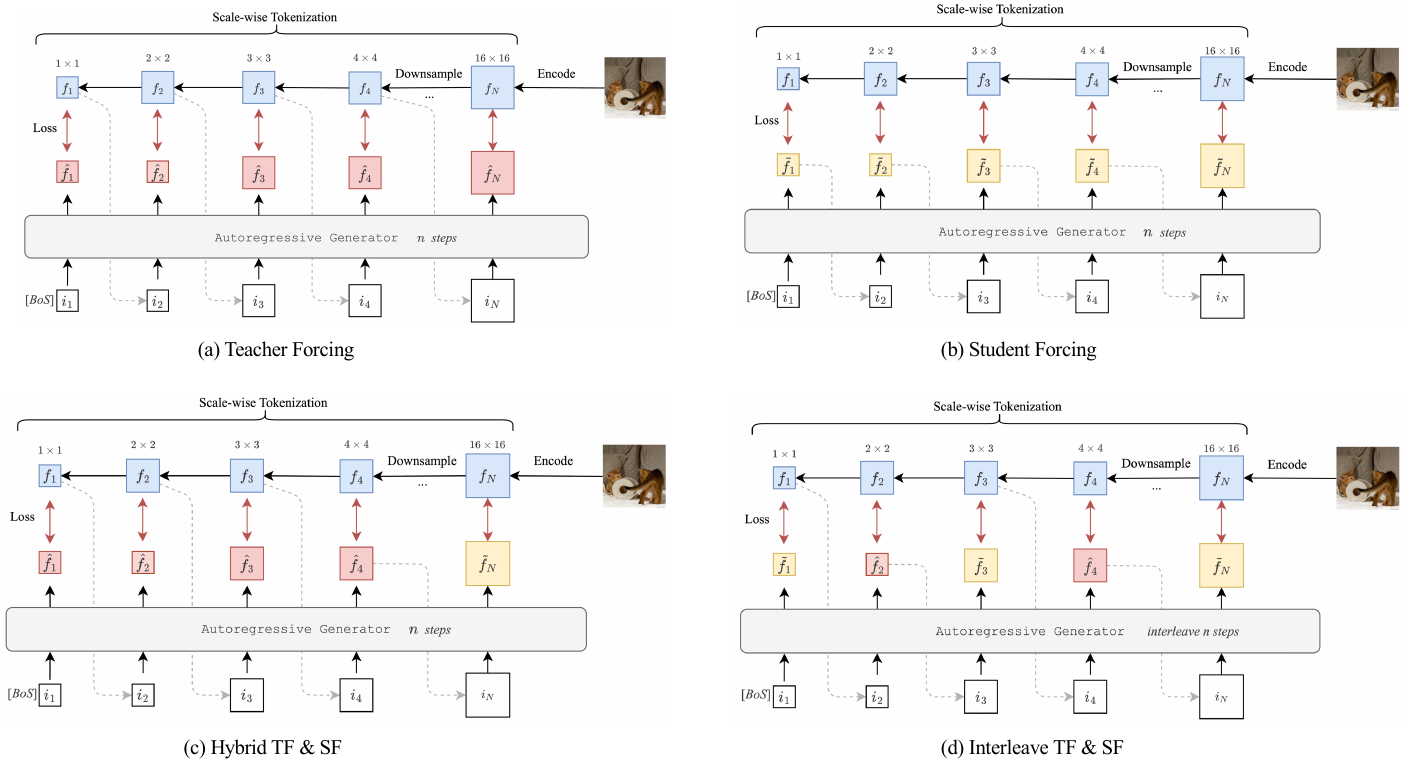}
\end{center}
\vspace{-1em}
\label{fig:sf}
\caption{
\textbf{Illustration of different student forcing training schemas}:
\protect\hypertarget{fig:sf:a}{}(a) Teacher Forcing (TF) uses ground-truth latents at all scales during training.
\protect\hypertarget{fig:sf:b}{}(b) Student Forcing (SF) uses predicted latents only, simulating test-time conditions.
\protect\hypertarget{fig:sf:c}{}(c) Hybrid TF~\&~SF applies teacher forcing at early scales and student forcing at later ones.
\protect\hypertarget{fig:sf:d}{}(d) Interleave TF~\&~SF alternates between teacher and student forcing across scales.
}

\vspace{-1em}
\end{figure*}

% ------------------------------------------------------------
\subsection{Naïve Integration of Student Forcing}
\label{sec:sf}
Formally, at each scale $i$, the model takes its own previously generated latents and predicts the next-scale latent:
\begin{equation}
    \tilde f_i = g_\theta(\tilde f_{1:i-1}), \qquad 
    \mathcal{L}_{\text{SF}} = \sum_{i=1}^{N} \ell(\tilde f_i, f_i).
\end{equation}

To investigate whether SF can close the train–test gap in VAR, we take a fully trained FlexVAR~\citep{jiao2025flexvar} checkpoint (100 epochs on Imagenet-256) and continue training for 20 epochs under four SF schedules designed to expose the model to different degrees of self-generated inputs:  

\begin{itemize}
    \item \emph{SF only (Fig.~\hyperlink{fig:sf:b}{3(b)})}: all scales are generated autoregressively; no ground-truth inputs are used during training.

    \item \emph{Alternate}: TF and SF updates alternate every iteration, providing a 1:1 mixture of GT and self-generated inputs.

    \item \emph{Interleave (Fig.~\hyperlink{fig:sf:c}{3(c)})}: TF is applied to odd-index scales while SF is applied to even-index scales, allowing mixed supervision within a single forward pass.

    \item \emph{Hybrid (Fig.~\hyperlink{fig:sf:d}{3(d)})}: TF is used up to a specified scale $k$, after which SF is applied to all finer scales, gradually transitioning from ground-truth to self-generated latents.
\end{itemize}

We present the result in Tab.~\ref{tab:sf_ablation}. Across all schedules, two consistent issues emerge.  
First, SF greatly increases training cost: autoregressive rollout removes parallelism across scales, making each iteration substantially slower than teacher-forcing.  
Second, directly train on self-generated latents destabilizes learning as early predictions largely drift away from the ground-truth targets, leading to misaligned intermediate supervision that is inapproarite for training. 
While the hybrid schedule mitigates the instability to some extent by delaying the introduction of SF, none of the schedules approach the stability or quality of teacher-forcing. The key finding is that naïve student-forcing, despite being conceptually aligned with autoregressive inference, fails to address the train–test gap. Instead, it introduces supervision inconsistency, amplifies noise across scales, and incurs high computational overhead, highlighting the need for a more principled approach.

% 4. SF
\begin{table}[t]
\footnotesize
\centering
\caption{\textbf{Comparison on student forcing schedules.} \textit{Alternate}, \textit{Interleave}, and \textit{Hybrid} involve a mixture with teacher forcing.}
\resizebox{0.7\columnwidth}{!}{
\begin{tabular}{lc|cccc}
\toprule
{Schedule} & {AR Model} & {FID$\downarrow$} & {IS$\uparrow$} & {Pre$\uparrow$} & {Res$\uparrow$} \\
\midrule
\rowcolor{gray!10}
Teacher Forcing           & FlexVAR-d16 & 3.83 & 248.82 & 0.83 &  0.48  \\
\midrule
\rowcolor{adobebg}
Student Forcing & & & & & \\
Full           & FlexVAR-d16 & 16.56 & 135.23 & 0.65 &  0.43  \\
Alternate                 & FlexVAR-d16 & 9.31 & 226.53 & 0.75 &  0.38  \\
Interleave                & FlexVAR-d16 & 9.47 & 399.66 & 0.86 &  0.32   \\
Hybird @ 8                & FlexVAR-d16 & 10.83 & 195.49 & 0.72 &  0.40   \\
Hybird @ 16               & FlexVAR-d16 & 5.90 & 286.87 & 0.83 & 0.41   \\
\bottomrule
\end{tabular}
}
\vspace{-3pt}

\label{tab:sf_ablation}
\end{table}

% These noisy states break the consistency required for effective coarse-to-fine modeling and result in degraded reconstruction quality.

% ------------------------------------------------------------
\subsection{SAR: A Stable Student-Forcing Strategy}
To address the aforementioned issues, we propose Self-autoregressive Refinement (\ours) for efficient and stable optimization at all scales. We introduce Stagger-Scale Rollout (SSR) to implement efficient student-forcing (only 2 rollout steps needed), and Contrastive Student-Forcing Loss (CSFL) to provide consistent training supervision.

% ------------------------------------------------------------
\subsubsection{Stagger-Scale Rollout (SSR)}

To make student-forcing practical for scale-wise autoregression, we introduce Stagger-Scale Rollout (SSR), a lightweight two-step rollout strategy designed to meet three requirements: 
(1) It must expose the model to its own predictions so that training more faithfully matches the autoregressive behavior used at inference.  
(2) The self-generated inputs must remain \emph{meaningful}; naïve SF quickly produces noisy, drifted latents that break the hierarchical structure and destabilize learning.  
(3) The rollout must remain computationally efficient and avoid the expensive full autoregressive unrolling that makes naïve SF prohibitively slow.
SSR addresses all three issues through a short, controlled rollout that preserves clean supervision while introducing limited but useful exposure to model-generated states.

The process begins with a standard teacher-forcing forward pass using ground-truth latent inputs. At each scale, the model predicts:
\begin{equation}
    \hat f^{(T)}_i = g_\theta(f_{1:i-1}),
\end{equation}
% producing model predictions under ground-truth inputs.

% Next, we construct a scale-shifted self-generated input. Each teacher-forced prediction is upsampled to the resolution expected at the next scale:
Such predictions can be upsampled and shifted to become the input to the next scale:
\begin{equation}
    \tilde f^{(T)}_i = \mathrm{Upsample}(\hat f^{(T)}_i),
\end{equation}
which follows the autoregressive inference.

We then perform a one-step parallelized student-forcing rollout by feeding the shifted latents into the model:
\begin{equation}
    \hat f^{(S)}_i = g_\theta(\tilde f^{(T)}_{1:i-1}),
\end{equation}
resulting in single-step student-forced outputs for all scales.

This staggered design yields two aligned trajectories: the teacher-forced predictions $\hat f^{(T)}_{1:N}$, and the one-step student-forced predictions $\hat f^{(S)}_{2:N}$.
By limiting the rollout to only two student-forcing steps, SSR provides just enough exposure to inference-like dynamics to reduce the train–test gap, while keeping the intermediate inputs structured and additional computational cost to only one extra NFE. Deeper rollouts, in contrast, accumulate noise rapidly and offer no additional benefit as observed in our empirical analysis (Tab.~\ref{tab:sf_ablation}).

\begin{figure*}[t]
\begin{center}
   \includegraphics[width=\textwidth]{figures/SSR.jpg}
\end{center}
\caption{
\textbf{Illustration of SAR.} The image is encoded into multi-scale latents $\{f_i\}$, which condition an autoregressive generator. In the first forward pass, the model performs teacher forcing and predicts $\hat f^{(T)}_i$ at all scales. These predictions are then upsampled to form scale-shifted inputs $\tilde f^{(T)}_i$, enabling a second forward pass that produces student-forced predictions $\hat f^{(S)}_i$. Teacher-forcing loss provides ground-truth supervision, while the contrastive student-forcing loss aligns student-forced outputs with their teacher-forced counterparts. Together, these two passes form the Stagger-Scale Rollout used in SAR.
}
\label{fig:sar}
\vspace{-3mm}
\end{figure*}

\begin{algorithm}[t]
\caption{Self-Autoregressive Refinement}
\label{alg:sar}
\begin{algorithmic}[1]

\State \textbf{Input:} image $x$, label $y$, scales $\{h_1,\dots,h_N\}$ 
\State \textbf{Modules:} multi-scale encoder $\mathcal{E}$, VAR generator $g_\theta$
\State \textbf{Hyperparameter:} SF weight $\gamma$

\State \textcolor{PineGreen}{\# Encode ground-truth latents for all scales}
\For{$i = 1$ to $N$}
    \State $f^{\text{GT}}_i = \mathcal{E}(x)$ resized to $(h_i \times h_i)$
\EndFor

\State \textcolor{RoyalBlue}{\# ======== Pass 1: Teacher Forcing (TF) ========}
\State \textcolor{PineGreen}{\# Construct inputs by upsampling GT latents $i \rightarrow i{+}1$}
\For{$i = 1$ to $N{-}1$}
    \State $u_i = \mathrm{Upsample}(f^{\text{GT}}_i \rightarrow h_{i+1})$
\EndFor

\State \textcolor{PineGreen}{\# Single TF forward pass across all scales}
\State $(\hat f^{(T)}_1,\dots,\hat f^{(T)}_N) = g_\theta(y,\{u_i\})$
\State $\mathcal{L}_{\mathrm{TF}} = \ell(\hat f^{(T)}_1,\dots,\hat f^{(T)}_N;\; f^{\text{GT}}_1,\dots,f^{\text{GT}}_N)$

\State \textcolor{PineGreen}{\# Sample TF predictions to build SF inputs}
\For{$i = 1$ to $N$}
    \State $\tilde f^{(T)}_i = \mathrm{Sample}(\hat f^{(T)}_i)$ \Comment{e.g., argmax}
\EndFor

\State \textcolor{RoyalBlue}{\# ===== Pass 2: One-Step Student Forcing (SF) =====}
\State \textcolor{PineGreen}{\# Scale-shift sampled prediction and run SF}
\State $u^{\mathrm{SF}} = \mathrm{Upsample}(\tilde f^{(T)}_k \rightarrow h_{k+1})$
\State $\hat f^{(S)}_{k+1} = g_\theta(y,\,u^{\mathrm{SF}})$
\State $\mathcal{L}_{\mathrm{CSF}} = \ell(\hat f^{(S)}_{k+1},\; f^{\text{GT}}_{k+1})$

\State \textcolor{RoyalBlue}{\# ========= Combine losses and update =========}
\State $\mathcal{L} = \mathcal{L}_{\mathrm{TF}} + \gamma \cdot \mathcal{L}_{\mathrm{CSF}}$
\State Backpropagate and update $\theta$

\end{algorithmic}
\end{algorithm}

\vspace{-2mm}

% ------------------------------------------------------------
\subsubsection{Contrastive Student-Forcing Loss (CSFL)}

While SSR provides a stable and efficient way to obtain both teacher-forced and student-forced trajectories, the key challenge remains: \emph{how should the student-forced predictions be supervised?}  
Directly comparing $\hat f^{(S)}_i$ to the ground-truth $f_i$ is ineffective, because student-forced latents may deviate from the teacher-forced hierarchy, especially at early scales. This mismatch creates ambiguous or conflicting supervision signals, which was precisely the failure case observed in naïve SF (Sec.~\ref{sec:sf}).

To address this issue, we introduce a Contrastive Student-Forcing Loss (CSFL) that aligns student-forced predictions with their teacher-forced counterparts instead of the ground-truth. The intuition is that the teacher-forced trajectory $\hat f^{(T)}_i$ provides a clean, stable estimate of the next-scale latent, and student-forced predictions should learn to remain consistent with this trajectory even when conditioned on self-generated inputs.

Formally, for each scale $i$, we compute two prediction paths:
\begin{align}
    \hat f^{(T)}_i &= g_\theta(f_{1:i-1}), \\
    \hat f^{(S)}_i &= g_\theta(\tilde f^{(T)}_{1:i-1}),
\end{align}
where $\hat f^{(T)}_i$ is the teacher-forced prediction and $\hat f^{(S)}_i$ is the one-step student-forced prediction obtained via SSR.

The total training loss consists of two terms.  
The first is the standard teacher-forcing loss:
\begin{equation}
    \mathcal{L}_{\text{TF}} = \sum_{i=1}^{N} \ell\big(\hat f^{(T)}_i, f_i\big),
\end{equation}
which provides ground-truth supervision and stabilizes the coarse-to-fine hierarchy.

The second term enforces consistency between teacher-forced and student-forced predictions:
\begin{equation}
    \mathcal{L}_{\text{CSF}} =
    \sum_{i=2}^{N} \ell\big(\hat f^{(S)}_i, \hat f^{(T)}_i\big),
\end{equation}
encouraging the model to produce latents that remain close to the teacher-forced trajectory even when using imperfect self-generated inputs.

Balancing these two components yields the full SAR training objective:
\begin{equation}
    \mathcal{L}_{\text{SAR}} 
    = \mathcal{L}_{\text{TF}}
    + \gamma \, \mathcal{L}_{\text{CSF}},
\end{equation}
where $\gamma$ controls the strength of student-forcing alignment.

Together with SSR, the CSFL provides \emph{structured, inference-aligned supervision}: the teacher-forced branch ensures stable training, and the student-forced branch learns to follow this stable trajectory even when conditioned on self-generated latents. This resolves the supervision inconsistency that destabilized naïve SF and forms the core of our SAR training strategy.

\section{Experiment}

\subsection{Implementation Details}

We train SAR on ImageNet-1K at 256$\times$256 resolution on three model scales following VAR~\citep{VAR}. SAR is formulated as a post-training refinement stage designed to further enhance generative quality after conventional teacher-forcing training.
We initialize SAR from sufficiently trained FlexVAR checkpoints and apply an additional 10 epochs of SAR training. To ensure a fair comparison with all baselines, we keep the total number of training epochs fixed. Specifically, for the three VAR scales, we resume from the 170/240/340-epoch FlexVAR checkpoints (trained with teacher forcing) and then train SAR for 10 additional epochs, matching the overall training epochs of competing methods.
We use the AdamW optimizer with $\beta_1 = 0.9$, $\beta_2 = 0.95$, and a weight decay of 0.05. The learning rate is set to 1e$^{-4}$ without warm-up. All experiments are conducted on NVIDIA A100 GPUs with 80GB of memory.

\subsection{State-of-the-Art Image Generation}
\begin{table*}[t]
\centering

\caption{
\textbf{Generative model comparison on class-conditional ImageNet 256$\times$256}.
Metrics include Fréchet inception distance (FID), inception score (IS), precision (Pre) and recall (Rec).
Step: the number of model runs needed to generate an image. 
Time: the relative inference time of VAR-$d30$~\cite{VAR}. 
We present models with a size $\leq$ 1B.
}\label{tab:main}

\setlength{\tabcolsep}{3.2mm}{}
\small
{
\begin{tabular}{l|cc|cc|cc|cc}
\toprule
 Model & Epochs & Param & Step & Time & FID$\downarrow$ & IS$\uparrow$ & Pre$\uparrow$ & Rec$\uparrow$ \\
\midrule
\rowcolor{gray!10}
\multicolumn{9}{c}{\emph{Generative Adversarial Networks (GAN)}} \\
 BigGAN~\citep{biggan}  & --  & 112M & 1 & $-$  & 6.95  & 224.5 & 0.89 & 0.38 \\
 GigaGAN~\citep{gigagan}  & --  & 569M & 1 & $-$  & 3.45  & 225.5 & 0.84 & 0.61 \\
 StyleGan-XL~\citep{stylegan-xl}  & --  & 166M & 1 & 0.2 & 2.30  & 265.1 & 0.78 & 0.53 \\
\midrule
\rowcolor{gray!10}
\multicolumn{9}{c}{\emph{Diffusion Models}} \\
 ADM-U~\citep{adm}       & 400  & 554M & 250  & 118 & 3.94  & 186.7 & 0.82 & 0.52 \\
 CDM~\citep{cdm}         & 2160  & $-$  & 8100 & $-$ & 4.88  & 158.7 & $-$  & $-$  \\
 LDM-4-G~\citep{ldm}     & 200  & 400M & 250  & $-$ & 3.60  & 247.7 & $-$  & $-$  \\
 DiT-L/2~\citep{dit}     & 1400  & 458M & 250  & 2   & 5.02  & 167.2 & 0.75 & 0.57 \\
 DiT-XL/2~\citep{dit}    & 1400  & 675M & 250  & 2   & 2.27  & 278.2 & 0.83 & 0.57 \\
\midrule
\rowcolor{gray!10}
\multicolumn{9}{c}{\emph{Random-scan Manner (Mask Prediction)}} \\
 MaskGIT~\citep{chang2022maskgit}     & 555  & 227M & 8 & 0.4 & 6.18  & 182.1 & 0.80 & 0.51 \\
 RCG (cond.)~\citep{li2024return}  & 400  & 502M & 20 & 1.4 & 3.49  & 215.5 & $-$  & $-$  \\
\midrule
\rowcolor{gray!10}
\multicolumn{9}{c}{\emph{Raster-scan Manner (Token-wise Autoregressive)}} \\
 VQGAN~\citep{esser2021taming} & 100  & 227M & 256 & 7 & 18.65 & 80.4  & 0.78 & 0.26 \\
 RQTran.~\citep{lee2022autoregressive} & 50  & 821M & 68 & $-$ & 13.11 & 119.3 & $-$  & $-$  \\
 LlamaGen-XL~\citep{llamagen} & 300  & 775M & 256 & 27 & 2.62 & 244.08 & 0.80 & 0.57 \\
 AiM~\citep{aim}   & 350  & 763M & 256 & 12 & 2.56 & 257.2  & 0.81 & 0.57 \\
\midrule
\rowcolor{gray!10}
\multicolumn{9}{c}{\emph{Scaling-scan Manner (Scale-wise Autoregressive)}} \\
VAR-$d16$~\citep{VAR}        & 180  & 310M & 10 & 0.2 & 3.55  & 280.4 & 0.84 & 0.51 \\
FlexVAR-$d16$~\citep{jiao2025flexvar}  & 180  & 310M & 10 & 0.2 & 3.05  & 291.3 & 0.83 & 0.52 \\
\rowcolor{adobebg}
\quad w/ \ours (Ours)        & + 10  & 310M & 10 & 0.2 & 2.89  & 266.6 & 0.81 & 0.55 \\
VAR-$d20$~\citep{VAR}        & 250  & 600M & 10 & 0.3 & 2.95  & 302.6 & 0.83 & 0.56 \\
FlexVAR-$d20$~\citep{jiao2025flexvar}  & 250  & 600M & 10 & 0.3 & 2.41  & 299.3 & 0.85 & 0.58 \\
\rowcolor{adobebg}
\quad w/ \ours (Ours)        & + 10  & 600M & 10 & 0.3 & 2.35  & 293.3 & 0.82 & 0.59 \\
 VAR-$d24$~\citep{VAR}       & 350  & 1.0B & 10 & 0.5 & 2.33  & 312.9 & 0.82 & 0.59 \\
FlexVAR-$d24$~\citep{jiao2025flexvar}  & 350  & 1.0B & 10 & 0.5 & 2.21  & 299.1 & 0.83 & 0.59 \\
\rowcolor{adobebg}
\quad w/ \ours (Ours)        & + 10  & 1.0B & 10 & 0.5 & 2.14  & 315.5 & 0.81 & 0.59 \\

\bottomrule
\end{tabular}

\vspace{-2mm}
}
\end{table*}

We compare our SAR-enhanced models with a broad range of state-of-the-art generative approaches on the ImageNet-1K benchmark, including GANs, diffusion models, random-scan and raster-scan autoregressive models, and the latest scaling-scan autoregressive methods. The quantitative results on 256$\times$256 resolution are summarized in Tab.~\ref{tab:main}.

\subsubsection{Overall comparison on ImageNet 256$\times$256.}
To ensure a fair comparison, we report results for all models with parameter sizes $\leq$1B. Across all three scales, applying our proposed SAR consistently improves the performance of the FlexVAR baselines, achieving state-of-the-art results among scaling-scan autoregressive models.
In particular, SAR yields clear FID improvements at every model size: 2.89 vs.\ 3.05 for the 310M model, 2.35 vs.\ 2.41 for the 600M model, and 2.14 vs.\ 2.21 for the 1B model. This demonstrates that SAR effectively enhances fine-scale generation quality while maintaining the efficient inference structure of scale-wise autoregression.

\subsubsection{Training Dynamics}

Figure~\ref{fig:teaser} (a) provides a joint visualization of the training behavior of \ours and its performance relative to state-of-the-art generative models. We compare three variants trained on ImageNet $256 \times 256$:

\begin{itemize}
    \item \textit{FlexVAR}~\citep{jiao2025flexvar}: a FlexVAR-d16 model trained from scratch for 180 epochs.
    \item \textit{SAR (from scratch)}: \ours trained from scratch for 180 epochs.
    \item \textit{SAR (init FlexVAR)}: \ours initialize from a well-trained FlexVAR model at 170 epochs, then further train for 10 epochs.
\end{itemize}

We observe several important trends:

\begin{enumerate}
    \item \textbf{Faster convergence.}  
    SAR trained from scratch descends sharply in the early stage and consistently maintains a lower FID across training. This indicates that \ours provides a stable student-forcing training scheme and could effectively improve teacher-forcing training.

    \item \textbf{Lower final error.}  
    By Epoch~180, SAR achieves a noticeably lower FID than FlexVAR. The smoother convergence curve suggests improved stability in token prediction during rollouts.

    \item \textbf{Strong initialization benefits.}  
    Initializing SAR from FlexVAR drastically accelerates early-stage optimization. Within only a few epochs, SAR quickly surpasses the best performance of a fully-trained FlexVAR model, converging to the lowest FID among all variants.

\end{enumerate}

Overall, the training analysis confirms that SAR not only enhances sample quality but also makes student-forcing training in scale-wise AR effective and stable.

\subsubsection{Model Comparison}

Figure~\ref{fig:teaser} (b) visualizes the throughput--FID trade-off across popular generative model families.

Key observations include:

\begin{enumerate}
    \item \textit{Diffusion models} (purple) generally achieve good FID but suffer from low throughput due to long sampling trajectories.
    \item \textit{Next-token AR models} (greens) improve throughput by using autoregression in latent space, but token-level causality limits their global coherence and negatively impacts FID.
    \item \textit{Next-scale AR models} (blues), including VAR and FlexVAR, provide an efficient coarse-to-fine sampler that greatly boosts throughput.
    \item \textit{\ours} (red) attains the \emph{best overall trade-off}:  
    the \textit{highest throughput among autoregressive models} and further imporve next-scale prediction AR model with the \textit{lowest FID across all AR baselines}.  
\end{enumerate}

The introduction of \ours retains the structural advantages of VAR while improving generation quality.

% ============================================
\subsection{Quanlitive Results}

\begin{figure*}[h]
\begin{center}
   \includegraphics[width=0.9\linewidth]{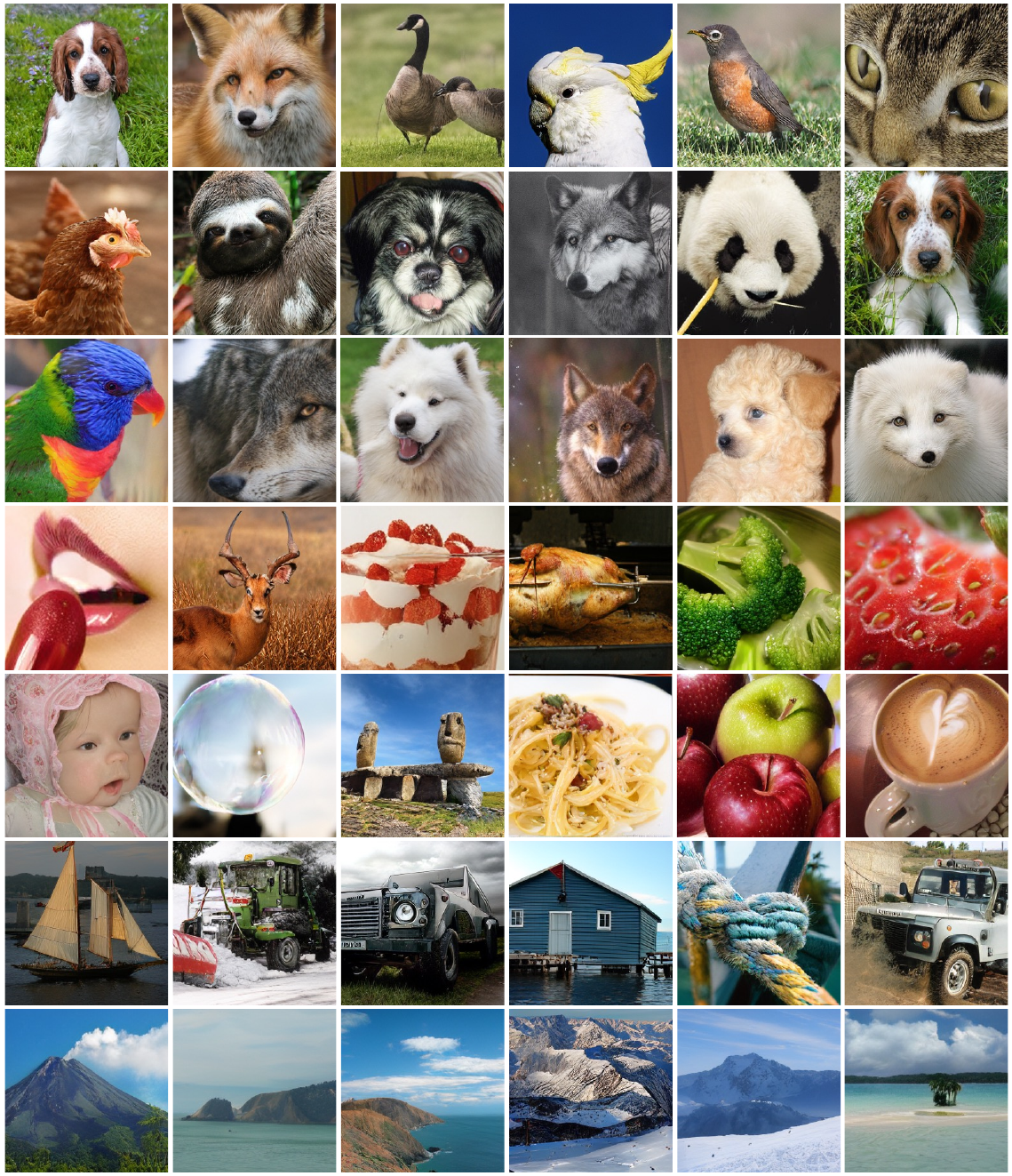}
\end{center}
   \caption{
    Images generated by \ours on ImageNet 256$\times$256.}
\label{fig:results}
\end{figure*}

We show the qualitative results of \ours in Figure~\ref{fig:results}.

\newpage

% ============================================
\subsubsection{Visualization of the Generation Process}

\begin{figure*}[t]
\begin{center}
   \includegraphics[width=0.95\linewidth]{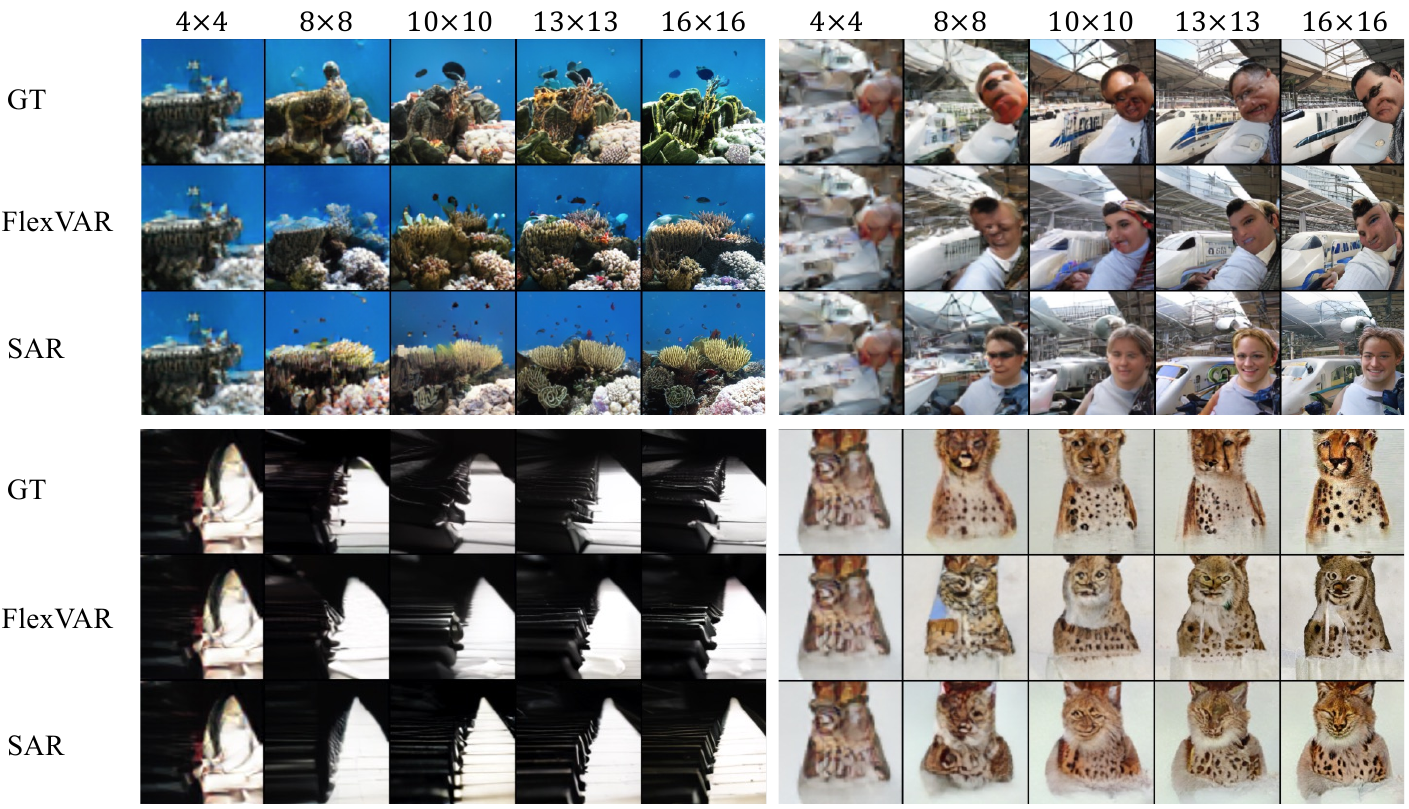}
\end{center}
\caption{
Comparison of the generation process of \ours and FlexVAR.  
Both models start from the same $4 \times 4$ latent and follow identical scale schedules.  
\ours demonstrates smoother transitions and stronger error correction across scales.}
\label{fig:gen_comp}
\end{figure*}

To better understand the qualitative differences between \ours and FlexVAR, we visualize the full generation trajectory of both models under identical initial conditions.  
We begin by selecting images from the ImageNet validation set, encoding them using the VAE to obtain the scale-wise latent representations. At inference time, both models are initialized with the same $4 \times 4$ latent, and perform coarse-to-fine autoregressive generation following the scale schedule:
\[
4 \rightarrow 8 \rightarrow 10 \rightarrow 13 \rightarrow 16.
\]

Figure~\ref{fig:gen_comp} presents the intermediate outputs at each scale for both \ours and FlexVAR.

Across all examples, two consistent patterns emerge:

\begin{enumerate}
    \item \textbf{Smooth and stable refinement.}  
    \ours exhibits smoother transitions between scales, with fewer abrupt changes in structure or texture.

    \item \textbf{Error correction during refinement.}  
    When early-scale predictions contain distortions or misaligned structures, \ours reliably corrects these errors at subsequent scales. In contrast, FlexVAR often amplifies early mistakes, leading to degraded high-resolution details.
\end{enumerate}

These visualizations highlight the student-forcing–aligned training in \ours produces more stable hierarchical latents, enabling more coherent and robust refinement during generation.

\subsection{Ablation Study}

\subsubsection{Effect of sampling strategy in student-forcing rollout.}

% \begin{table}[t]
\begin{wraptable}{r}{0.55\linewidth}
\footnotesize
\centering
\caption{\textbf{Ablation on Student Forcing sampling methods.} Comparison of different SF sampling strategies during training.}

\resizebox{\linewidth}{!}{
\begin{tabular}{c|c|cccc}
\toprule
{Sampling} & {CFG} & {FID$\downarrow$} & {IS$\uparrow$} & {Pre$\uparrow$} & {Res$\uparrow$} \\
\midrule
\rowcolor{gray!10}
\multicolumn{2}{c|}{Baseline}  & 3.47 & 325.76 & 0.84 &  0.50  \\
\midrule
\xmark & \xmark & 2.98 & 259.93 & 0.81 &  0.55  \\
\cmark & \xmark & 2.94 & 278.69 & 0.82 &  0.54  \\
\cmark & \cmark & 2.89 & 266.64 & 0.81 & 0.55 \\
\bottomrule
\end{tabular}
}

\label{tab:sampling_ablation}
\vspace{-5mm}
\end{wraptable}
% \end{table}

A key component of SAR is the sampling procedure used to generate student predictions during the rollout stage. We therefore ablate the sampling strategy used in SAR training and study its impact on final image quality.

We evaluate three variants:  
(1) \emph{No sampling}, where we directly use the argmax token as the student prediction;  
(2) \emph{Stochastic sampling}, where tokens are sampled from the predictive distribution;  
(3) \emph{Sampling with classifier-free guidance (CFG)}.  
For all sampling-based variants, we use top-$k{=}900$ and top-$p{=}0.95$. For CFG sampling, we set the guidance scale to 2.5. Results are reported in Tab.~\ref{tab:sf_ablation}.

The results show that incorporating stochasticity improves FID and recall over deterministic argmax, while CFG further enhances image fidelity and overall generation quality. Notably, the CFG sampling variant achieves the best performance with an FID of 2.89.

% ============================================
\clearpage
\begin{figure}[ht]
\begin{center}
   \includegraphics[width=0.99\linewidth]{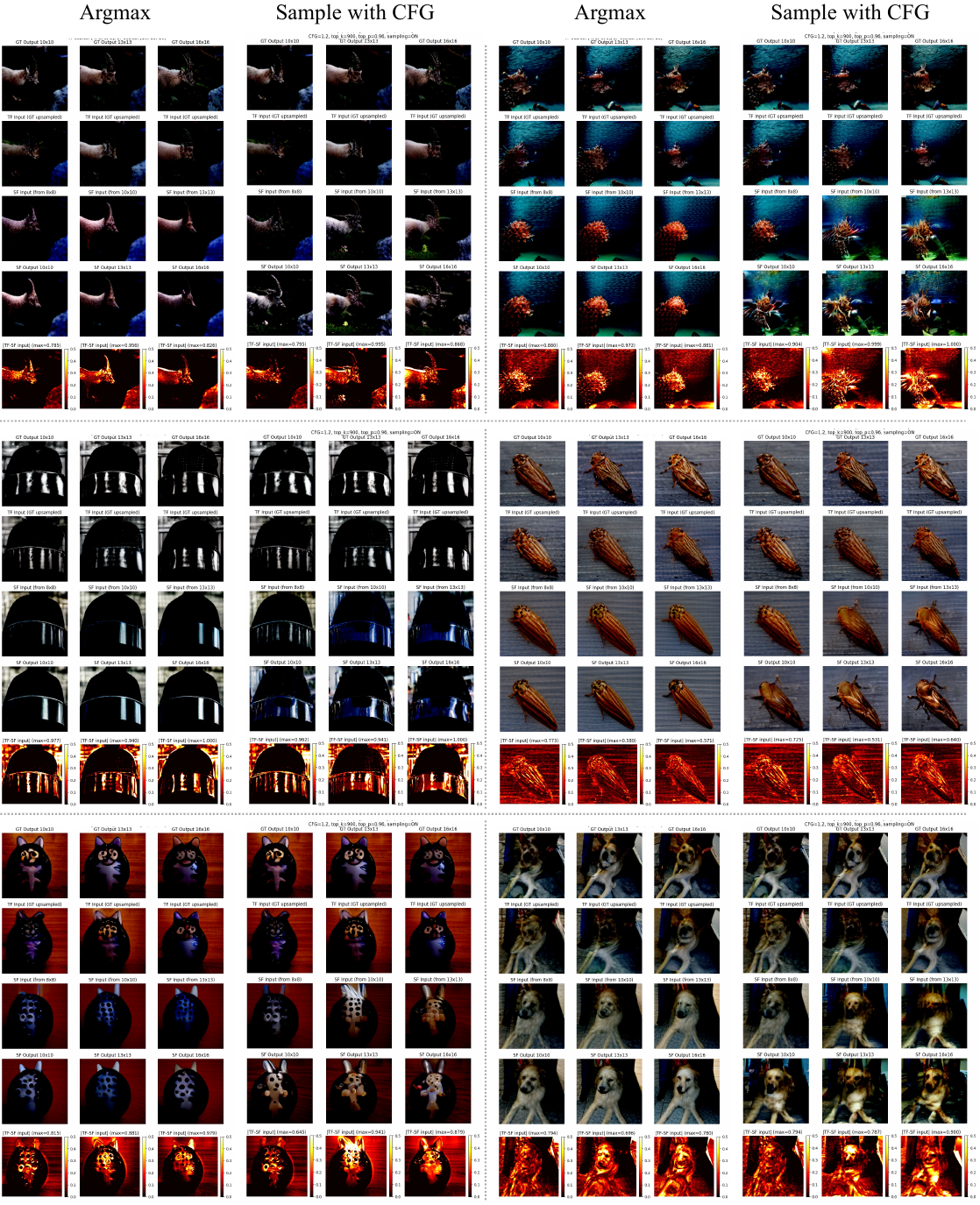}
\end{center}
    \caption{
    \textbf{Visualization of student-forcing inputs during training.}  
    We show ground-truth latents, teacher-forcing (TF) inputs, student-forcing (SF) inputs, SF predictions, and the corresponding difference maps across two consecutive rollout steps ($10\!\rightarrow\!13$ and $13\!\rightarrow\!16$).
    }
\label{fig:sf-vis}
\end{figure}

\subsubsection{Visualization of Student-Forcing Inputs During Training}
\label{sec:D}

To better understand the behavior of student-forcing (SF) within our SAR training framework, we visualize the intermediate latents produced during two consecutive SF rollout steps. This provides an intuitive, fine-grained view of how self-generated latents evolve, how errors propagate across scales, and how they differ from the teacher-forcing (TF) pathway.

Our visualization uses the same notation and formulation introduced in Sec.~\ref{sec:sf}. Recall that naïve student-forcing predicts the next-scale latent from its own previously generated latents:
\begin{equation}
    \tilde f_i = g_\theta(\tilde f_{1:i-1}), \qquad
    \mathcal{L}_{\mathrm{SF}} = \sum_{i=1}^{N} \ell(\tilde f_i, f_i),
\end{equation}
which exposes the model to inference-like conditions but suffers from drift, error amplification, and destabilized supervision.

In contrast, SAR replaces naïve SF with Stagger-Scale Rollout (SSR), which produces both teacher-forced predictions
\begin{equation}
    \hat f^{(T)}_i = g_\theta(f_{1:i-1}),
\end{equation}
and a one-step, structure-preserving student-forced trajectory:
\begin{align}
    \tilde f^{(T)}_i &= \mathrm{Upsample}(\hat f^{(T)}_i), \\
    \hat f^{(S)}_i &= g_\theta(\tilde f^{(T)}_{1:i-1}).
\end{align}

% ------------------------------------------------------------
\noindent\textbf{Visualization Setup.} 
Figure~\ref{fig:sf-vis} shows the intermediate latents obtained when performing two consecutive student forcing rollout steps, visualized across scale transitions:
$10 \rightarrow 13, 13 \rightarrow 16$.

Each row in the figure corresponds to a specific component in the SAR rollout pipeline:

\begin{itemize}
    \item \textit{Row 1: Ground-truth latents ($f_{10}$, $f_{13}$, $f_{16}$).}  
    These represent the teacher-forcing targets at each scale.

    \item \textit{Row 2: TF model inputs ($\mathrm{up}(f_{8})$, $\mathrm{up}(f_{10})$, $\mathrm{up}(f_{13})$).}  
    Ground-truth latents are upsampled to match the next-scale resolution.

    \item \textit{Row 3: SF inputs ($\tilde f^{(T)}_{8}$, $\tilde f^{(S)}_{10}$, $\tilde f^{(S)}_{13}$).}  
    These are obtained by shifting TF predictions to the next scale at scale 8:
    \[
        \tilde f^{(T)}_i = \mathrm{Upsample}(\hat f^{(T)}_i).
    \]

    And two consecutive SF steps at scales 10 and 13.

    \item \textit{Row 4: SF predictions ($\hat f^{(S)}_{10}$, $\hat f^{(S)}_{13}$, $\hat f^{(S)}_{16}$).}  
    These are produced by feeding the shifted predictions into the model:
    \[
        \hat f^{(S)}_i = g_\theta(\tilde f _{1:i-1}).
    \]

    \item \textit{Row 5: Difference maps ($\Delta_i$).}  
    To quantify the student-forcing deviation at each scale, we compute:
    \[
        \Delta_i = \left|\, f_i - \tilde f^{(S)}_i \,\right|.
    \]
\end{itemize}

Rows 3--5 make the effect of student-forcing explicit: how student-generated latents enter the next-scale prediction, how the model responds, and where errors emerge.

% ------------------------------------------------------------
\noindent\textbf{Observations.} 
Three consistent observations emerge from the visualization:

\begin{enumerate}
    \item \textbf{Error amplification across scales.}  
    Deviations introduced at coarse scales (e.g., $10 \rightarrow 13$) propagate and typically enlarge at finer scales ($13 \rightarrow 16$), demonstrating why naïve SF destabilizes training.

    \item \textbf{Structured corrections under SSR.}  
    Because SSR starts from teacher-forced predictions, the SF inputs remain structurally meaningful. As a result, the model corrects local distortions rather than diverging, which stabilizes hierarchical learning.

    \item \textbf{Alignment between trajectories.}  
    The CSFL drives $\hat f^{(S)}_i$ to match the stable TF predictions $\hat f^{(T)}_i$, yielding visibly smaller difference maps compared to naïve SF. This explains why SAR can leverage student-forcing without drifting off-manifold.
\end{enumerate}

Together, these visualizations clarify why naïve SF fails and why SAR succeeds: SSR ensures that self-generated inputs remain structured, while CSFL ensures that student-forced latents follow the correct predictive trajectory. This resolves the supervision inconsistency and the train–test gap in scale-wise autoregression.
\section{Conclusion}
We introduced Self-Autoregressive Refinement (SAR), a lightweight post-training framework that mitigates the train–test mismatch and scale-wise learning imbalance in previous Visual Autoregressive models. By incorporating Stagger-Scale Rollout for efficient exposure to self-generated latents and a Contrastive Student-Forcing Loss for stable refinement, SAR aligns training dynamics with inference behavior while adding minimal computational cost. Applied to state-of-the-art AR models, SAR consistently improves perceptual quality and robustness across model sizes, achieving up to 5.2\% FID reduction on ImageNet-256 with less than 5.5\% of pretraining compute. Our results demonstrate that fine-grained autoregressive refinement is both practical and effective, offering a scalable path toward stronger and more reliable visual autoregressive generation.

\clearpage
\newpage
\bibliographystyle{assets/plainnat}
\bibliography{main}

\end{document}